\documentclass[sigconf]{acmart}

\usepackage{booktabs} 

\usepackage{graphicx}
\graphicspath{{./figures/}}
\usepackage{subcaption}
\usepackage[ruled]{algorithm2e}
\usepackage{caption}
\usepackage{color}
\usepackage{ctable}
\usepackage{siunitx}
\usepackage{amsmath}
\usepackage{enumitem}
\usepackage{multirow}
\usepackage{balance}

\setcopyright{none}

\acmDOI{}

\acmISBN{}

\acmConference[DAC'19]{Design Automation Conference}{June 2--6, 2019}{Las Vegas, NV}
\acmYear{}
\copyrightyear{}

\newcommand{\citeall}[1]{\citeauthor{#1}~\citeyear{#1}~\cite{#1}}

\begin{document}

\title{Fast and Efficient Information Transmission with Burst Spikes \\
in Deep Spiking Neural Networks}


\author{Seongsik Park, Seijoon Kim, Hyeokjun Choe, Sungroh Yoon}
\affiliation{%
  \institution{Department of Electrical and Computer Engineering, Seoul National University}
}
\email{sryoon@snu.ac.kr}



\pagestyle{empty}

\begin{abstract}
The spiking neural networks (SNNs) are considered as one of the most promising artificial neural networks due to their energy-efficient computing capability.
Recently, conversion of a trained deep neural network to an SNN has improved the accuracy of deep SNNs.
However, most of the previous studies have not achieved satisfactory results in terms of inference speed and energy efficiency.
In this paper, we propose a fast and energy-efficient information transmission method with burst spikes and hybrid neural coding scheme in deep SNNs.
Our experimental results showed the proposed methods can improve inference energy efficiency and shorten the latency.
\end{abstract}



\maketitle

\section{Introduction}

With the recent development of deep learning, deep neural networks (DNNs) have shown outstanding performance in various applications.
DNNs, however, demand high computational power which raises concerns for high energy consumption when deployed in mobile environment.
To overcome this challenge, extensive studies have focused mainly on two approaches; reducing size of models~\cite{han2015deep,park2018quantized}, and designing application specific accelerators~\cite{han2016eie,park2018streaming}.

Although these efforts have been able to reduce energy consumption of DNNs, they have failed to fully take into account the efficiency and capability of information processing in the human brain.
To bridge the gap between the human brain and artificial neural networks (ANNs), spiking neural networks (SNNs) have been studied as the 3rd generation ANNs~\cite{maass1997networks}.


SNNs have potentials to process information rapidly and efficiently.
Unlike DNNs, neurons in SNNs can transmit information through a spike train, which is a series of spikes that contains temporal information.
In addition, a neuron integrates all the information received from synapses, and then generates a spike if the integrated value exceed a certain threshold.
This integrate-and-fire feature allows SNNs to exploit event-driven computation, which leads to much higher computational efficiency in neuromorphic architectures~\cite{merolla2014million,furber2014spinnaker}.


Despite their possibilities, SNNs have not shown comparable results with DNNs in various tasks.
One of the major reasons for SNN's limited applicability is lack of scalable training algorithms.
It is possible to train shallow SNNs consisting of two or three layers, using spiking-timing dependent plasticity (STDP) or error back-propagation, which are well-known training algorithm for SNNs and DNNs, respectively~\cite{diehl2015unsupervised,lee2016training}.
However, there have been few studies regarding training deep SNNs, such as VGG or ResNet.

As an alternative way of directly training deep SNNs, several studies have proposed methods of converting trained DNNs to deep SNNs~\cite{cao2015spiking,diehl2015fast,rueckauer2016theory,rueckauer2017conversion,kim2018deep}.
These approaches aim at energy-efficient inference in deep SNNs by adopting well-trained weights in DNNs.
Through these methods, the SNNs have shown comparable inference accuracy with that of DNNs in various tasks, especially image classification tasks.

Most of the related studies have focused on reducing the accuracy degradation caused by the conversion of DNNs to SNNs.
A few studies have considered the latency and efficiency of inference in SNNs~\cite{kim2018deep,park2016efficient,diehl2015fast}.
In addition, to the best of our knowledge, deep SNNs~\cite{kim2018deep} is the only relevant study utilizing the temporal information of spike trains in deep SNNs.
Their method was successful in delivering information quickly, but the large number of spikes lowered the efficiency of inference.

In this paper, we propose a fast and energy-efficient method to transmit information between subsequent neurons in deep SNNs.
The proposed method consists of a neural model for generating burst spikes and a layer-wise hybrid neural coding scheme.
We were inspired by the neuroscience research that burst spikes enhance the efficiency of the information transmission~\cite{connors1990intrinsic,izhikevich2003bursts,lisman1997bursts}, and the human brain generated different spike patterns depending on the role of the neurons~\cite{mochizuki2016similarity}.
We analyzed the effect of the proposed hybrid neural coding scheme through firing pattern analysis.

By using our method, we can improve the accuracy and efficiency of the deep SNNs in terms of latency and number of spikes in image classification tasks.
The contributions of this paper can be summarized as follows:
\begin{itemize}[topsep=0pt,itemsep=0ex,partopsep=1ex,parsep=1ex,leftmargin=*]
    \item We propose a spiking neuron model that exploits burst spikes to improve the efficiency of information transmission between subsequent neurons in deep SNNs.
    This is the first attempt to apply the concept of the burst spiking to deep SNNs.
    \item We suggest a layer-wise hybrid neural coding scheme depending on the functions of different types of layer.
    \item We demonstrate the effects of the hybrid neural coding scheme through analysis of spike patterns in deep SNN.
\end{itemize}

\section{Background and Related Work}
\subsection{Spiking Neural Networks}
\begin{figure*}[tbp]
    \centering
    \includegraphics[width=1.0\linewidth]{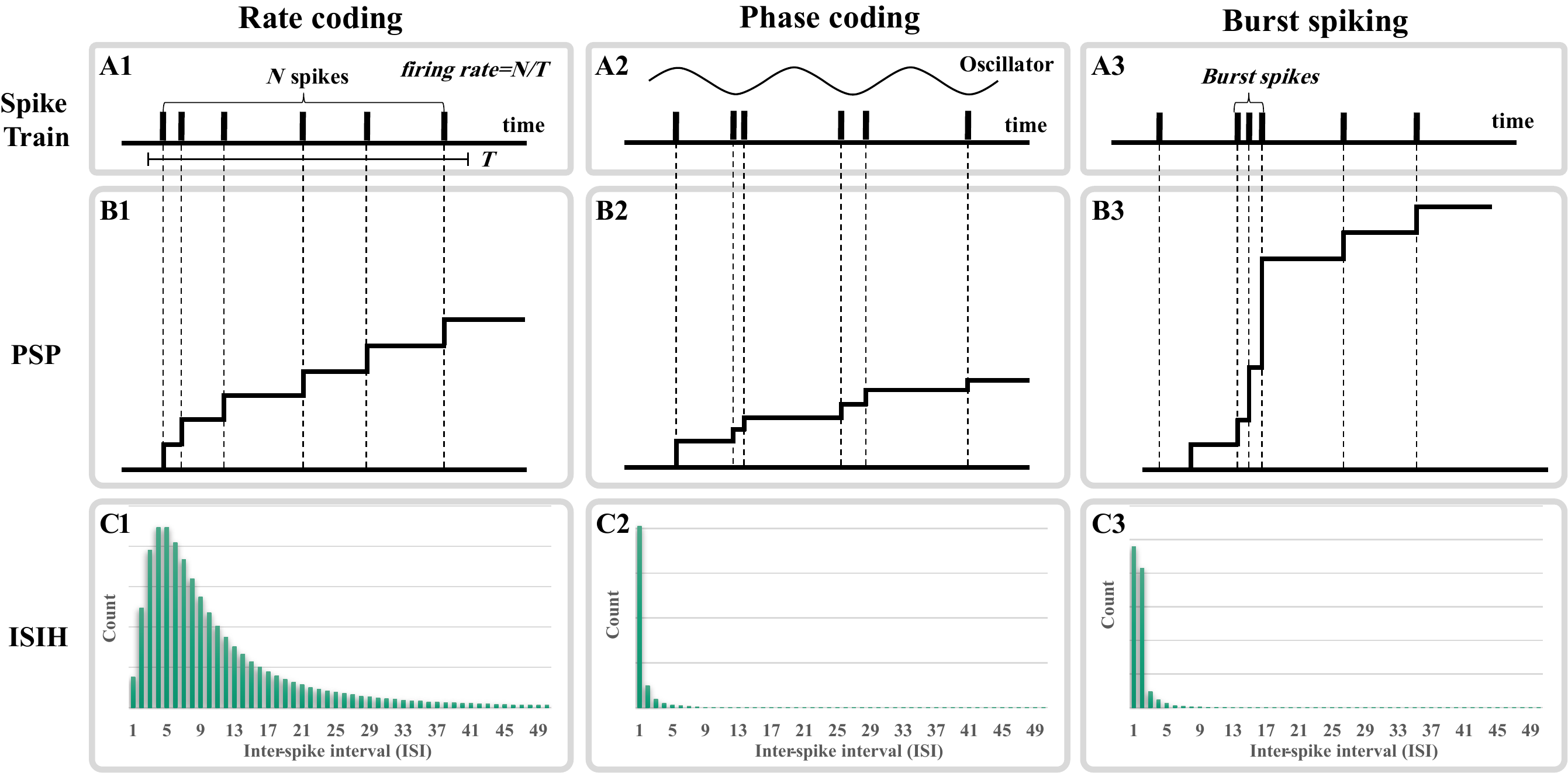}
	\vspace{-1.8em}
	\caption{(A) Spike train, (B) post-synaptic potentiation (PSP), and (C) inter-spike interval histogram (ISIH) of the integrate-and-fire (IF) neurons depending on the various types of neural coding}
	\label{fig:neural_coding}
	\vspace{-1.8em}
\end{figure*}

SNNs are composed of neurons which generate spike trains and synapses which allow a neuron to transmit spike trains to other neurons.
An integrate-and-fire (IF) neuron model, which is one of the most widely used models in SNNs, integrates post-synaptic potential (PSP) into the membrane potential $V_{\textrm{mem}}$.
The integrated PSP of the $j$th neuron in the $l$th layer at time $t$ is described as 
\begin{equation}
\label{eq:psp}
z_{j}^{l}(t) = \sum_{i}{w_{ij}^{l}\Theta_{i}^{l\textrm{-}1}(t)+b_{j}^{l}} \textrm{,}
\end{equation}
where $z_{j}^{l}(t)$ is the sum of PSPs, $w_{ij}^{l}$ is the synapse weight between $i$th neuron in $l\textrm{-}1$th layer, $\Theta_{i}^{l\textrm{-}1}(t)$ is a spike in the $i$th neuron in $l\textrm{-}1$th layer, and $b_{j}^{l}$ is the bias.

If the membrane potential exceeds a certain threshold $V_{\textrm{th}}$, the neuron generates a spike.
The spike generating process of the $i$th neuron in the $l$th layer at time $t$ can be formulated as follows:
\begin{equation}
\label{eq:spiking_func}
\Theta_{i}^{l}(t) = U(V_{\textrm{mem},i}^{l}(t-1) + z_{i}^{l}(t) - V_{\textrm{th}}(t)) \textrm{,}
\end{equation}
where $U(x)$ is a unit step function for representing the output spike.
If an IF neuron fires a spike, then the membrane potential is set to the resting potential $V_{\textrm{rest}}$, otherwise the potential is maintained.
This dynamics of the membrane potential is described as
\begin{equation}
\label{eq:vmem}
    V_{\textrm{mem},j}^{l}(t) =
    \begin{cases}
        V_{\textrm{rest},j}^{l}(t)  & \textrm{if}~\Theta_{j}^{l}(t)=1 \\
        V_{\textrm{mem},j}^{l}(t-1) + z_{j}^{l}(t)  & \
        \textrm{otherwise} \textrm{.}
    \end{cases}
\end{equation}

In SNNs, information is transmitted through spike train and processed only when spike occurs.
This event-driven feature of SNNs can improve computational energy-efficiency of neuromorphic architecture~\cite{merolla2014million,furber2014spinnaker}.
Despite the efficient processing possibility, SNNs have not yet been widely applied due to the difficulty in training.

\subsection{Neural Coding}
Neural coding is a method of representing information as spike trains. 
Various types of neural coding have been extensively studied, with rate and temporal coding~\cite{adrian1926impulses,gautrais1998rate,thorpe2001spike,thorpe1990spike,kayser2009spike}.

Rate coding is based on spike firing rate and has been widely used in several studies ~\cite{diehl2015fast,rueckauer2017conversion,rueckauer2016theory}.
It relies solely on the number of spikes over a certain time period (Fig.~\ref{fig:neural_coding}-A1), and induces the same size of PSP on each spike as shown in Fig.~\ref{fig:neural_coding}-B1.
For example, to represent 8 bits, $2^{8} = 256$ time steps are required. 
Thus, it is inefficient in terms of information capacity and transmission speed~\cite{gautrais1998rate}, and leads to a higher latency and energy consumption in deep SNNs.

Temporal coding is based on the temporal information in spike trains and includes time-to-first-spike~\cite{thorpe2001spike}, rank-order coding~\cite{thorpe1990spike}, and phase coding~\cite{kayser2009spike}. 
It has been proven to be effective in information transmission. However, to the best our knowledge, phase coding has only been successfully applied to deep SNNs~\cite{kim2018deep}.
Phase coding uses a periodic oscillation function as a global reference (Fig.~\ref{fig:neural_coding}-A2).
It produces PSP according to the phase of the oscillation function (Fig.~\ref{fig:neural_coding}-B2).

Burst spiking is a diverse spike pattern that contains a group of short inter-spike interval (ISI) spikes as shown in Fig.~\ref{fig:neural_coding}-A3.
Typically, burst spikes are known to be more reliable and carries more information than other spike trains. 
For example, it can dynamically induce synaptic potentiation which can improve information transmission efficiency as described in Fig.~\ref{fig:neural_coding}-B3.
Given this context, burst spikes can be applied to deep SNN to improve energy-efficiency of inference and reduce inference time.

\subsection{Conversion Methods of DNNs to SNNs}




The SNNs directly trained by STDP or error back-propagation have not shown comparable results to those of DNNs~\cite{lee2016training,diehl2015unsupervised}.
To take advantage of the energy-efficient information processing capability of the deep SNNs, DNN to SNN conversion methods have been studied recently~\cite{cao2015spiking,diehl2015fast,rueckauer2016theory,rueckauer2017conversion,kim2018deep}. 
The conversion method is based on a concept that imports weights of trained DNNs into the SNNs.

As an early stage of the research, to minimize accuracy degradation of the converted SNN, \cite{cao2015spiking} proposed a constrained model, removed biases and used average-pooling instead of max-pooling.
The authors then retrained the constrained model and applied retrained weights to SNN.

\begin{figure}[tbp]
    \centering
    \includegraphics[width=1.0\linewidth]{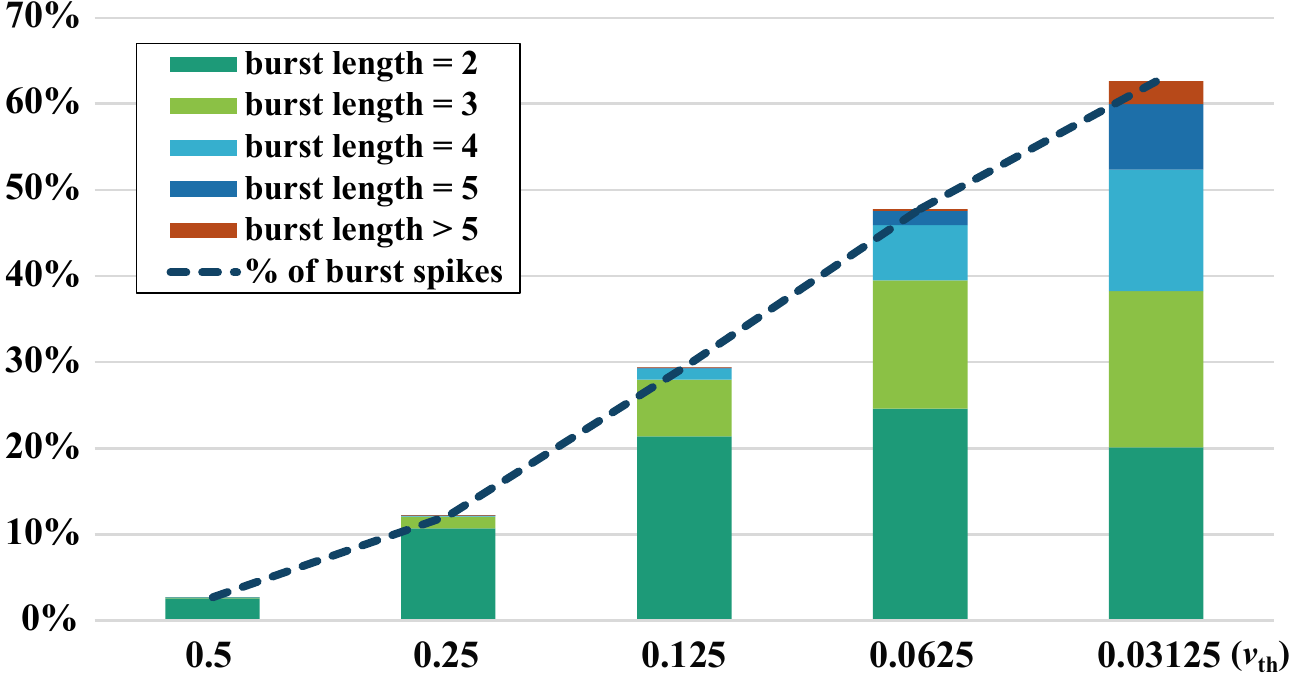}
	\vspace{-2.0em}
	\caption{Percentage of burst spikes for various $v_{\textrm{th}}$ as shown in Eq. \ref{eq:burst_vth}, and their composition by different burst lengths}
	\label{fig:burst_spike_rate}
	\vspace{-1.5em}
\end{figure}

Rate coding was introduced to deep SNNs to improve inference accuracy and time~\cite{diehl2015fast}.
The authors approximated DNN's activation value to SNN's firing rate through ReLU activation and data-based weight normalization.
This approach showed improvement in inference speed and accuracy for the MNIST dataset, however, it is only applicable to constrained models. 

To address such issues, normalized bias and spiking max pooling were proposed~\cite{rueckauer2016theory,rueckauer2017conversion}.
In addition, they proposed outlier robust normalization method.
They used reset-by-subtraction approach for preventing performance degradation caused by reset function of membrane potential.
The resting potential $V_{\textrm{rest}}$ is stated as
\begin{equation}
\label{eq:dnn2snn_vrest_rc}
    V_{\textrm{rest},j}^{l}(t) = V_{\textrm{mem},j}^{l}(t-1) + z_{j}^{l}(t) - V_{\textrm{th},j}^{l}(t)\Theta_{j}^{l}(t) \textrm{,}
\end{equation}
and the sum of PSPs to reduce information loss between neurons is represented as
\begin{equation}
\label{eq:dnn2snn_psp_rc}
z_{j}^{l}(t) = \sum_{i}{w_{ij}^{l}V_{\textrm{th},j}^{l}(t)\Theta_{i}^{l\textrm{-}1}(t)+b_{j}^{l}} \textrm{.}
\end{equation}
The results showed similar accuracy to that of DNN for on MNIST and CIFAR-10 datasets.
However, their approach of using rate coding may not fully exploit the temporal information in spike trains.

A recent study emphasized inference efficiency of deep SNNs~\cite{kim2018deep}.
The authors proposed weighted spike to utilize temporal information in spike trains.
We will refer to this method as \textit{phase coding} in the rest of this paper.
The oscillation function of their phase coding is given by 
\begin{equation}
\label{eq:dnn2snn_phase_ws}
    \Pi(t) = 2^{-(1+mod(t,k))} \textrm{,}
\end{equation}
where k is a period of the phase coding.
They used the PSP function shown in Eq.~\ref{eq:dnn2snn_psp_rc} and adjusted $V_{\textrm{th}}(t)$ according to the oscillation function as 
\begin{equation}
\label{eq:dnn2snn_vth_ws}
    V_{\textrm{th},i}^{l}(t) = \Pi(t)v_{\textrm{th}} \textrm{,}
\end{equation}
where $v_{\textrm{th}}$ is a threshold constant, which induces changes in effective weight 
$w_{ij}^{l}V_{\textrm{th},j}^{l}(t)$ 
depending on the phase as shown in Eq.~\ref{eq:dnn2snn_psp_rc}.
As a result, this method was able to perform image classification with lower latency than other deep SNN models with rate coding.

However, the transmission amount of information is bounded and more spikes are generated, because their method cannot change the transmission amount adaptively.
This can cause inefficiency in hidden layers.
Thus, there is still a need for novel inference methods that can reduce latency and number of spikes.

\section{Proposed Methods}
\subsection{Neural Model for Burst Spikes}
\ctable[
pos = t,
center,
caption = {Experimental results on CIFAR-10 with various configurations of neural coding in input and hidden layers (model: VGG-16, test accuracy of DNN model: 91.41\%)},
captionskip = -3ex,
mincapwidth = \columnwidth,
label = {tab:exp_result_nc},
doinside = {\small \def\arraystretch{0.7}}
]{llcrr}{
}{
    \toprule
    Input & Hidden & Accuracy (\%) & Latency & \# of spikes ($10^6$)\\
	\midrule
	real & rate & 91.06 & 1,500 & \textbf{9.334} \\
	real & phase & 90.91 & 1,500 & 39.301 \\
	real & burst & \textbf{91.34} & 1,500 & 49.830 \\
	\midrule
	rate & rate & 82.90 & 1,500 & \textbf{9.186} \\
	rate & phase & 36.39 & 1,500 & 78.238 \\
	rate & burst & \textbf{83.71} & 1,500 & 41.451 \\
	\midrule
	phase & rate & 82.38 & 1,500 & \textbf{1.164} \\
	phase & phase & 91.21 & 1,500 & 35.196 \\
	phase & burst & \textbf{91.41} & \textbf{1,125} & 6.920 \\
	\bottomrule   
	\vspace{-2.5em}
}

%

As mentioned before, the rate and phase coding lack efficiency in transmitting information such that they do not fit well in the deep SNNs.
In this paper, we propose an approach to achieving fast and energy-efficient inference in deep SNNs with burst spikes.
Our proposed method is motivated by the fact that burst spikes are adequate for transmitting information efficiently~\cite{connors1990intrinsic,izhikevich2003bursts,lisman1997bursts}.
To implement the functionality of burst spiking, we define the burst function as follows:
\begin{equation}
\label{eq:burst_function}
    g_{i}^{l}(t) =
    \begin{cases}
        \beta g_{i}^{l}(t-1) & \textrm{if}~\Theta_{i}^{l}(t-1)=1 \\
        1.0 & \textrm{otherwise} \ \textrm{,}
    \end{cases}
\end{equation}
where $\beta$ is a burst constant.
To obtain the effect of burst spikes, we applied the burst function to Eq.~\ref{eq:dnn2snn_psp_rc} through threshold adaptation which is depicted as 
\begin{equation}
\label{eq:burst_vth}
    V_{\textrm{th},i}^{l}(t) = g_{i}^{l}(t)v_{\textrm{th}} \textrm{.}
\end{equation}

The burst spikes induce synaptic potentiation (strengthening of synapse), which improves transmitting efficiency~\cite{krahe2004burst}.
Through Eqs.~\ref{eq:dnn2snn_psp_rc} and \ref{eq:burst_vth}, we can determine effective weights $\hat{w}_{ij}^{l}(t)$ due to burst spikes as 
\begin{equation}
\label{eq:burst_synaptic_potentiation}
\hat{w}_{ij}^{l}(t) = w_{ij}^{l}g_{i}^{l}(t) \textrm{,}
\end{equation}
and the proposed method can change the effective weights dynamically depending on the amount to transmit.

A burst is composed of a group of short-ISI spikes.
Thus, we can verify the occurrence of the burst spikes indirectly through the ISI histogram (ISIH).
In Fig.~\ref{fig:neural_coding}, C1, C2, and C3 illustrate the ISIH of rate coding, phase coding \cite{kim2018deep}, and proposed method for generating burst spikes, respectively.
As presented in Fig.~\ref{fig:neural_coding}-C1 (rate coding) and C3 (proposed method), the ratio of short-ISI spikes was considerably increased when the proposed method was applied.
In the case of phase coding (Fig.~\ref{fig:neural_coding}-C2), the ratio of short-ISI spikes was higher than that of the proposed method.

The proposed approach can control the precision of the information being transmitted according to the value of $v_{\textrm{th}}$.
Fig.~\ref{fig:burst_spike_rate} shows the total number of burst spikes and their composition by different burst lengths as $v_\textrm{th}$ varies.
We observed that, as the $v_\textrm{th}$ decreases, the portion of burst spikes increases in the whole spike trains, and longer bursts occur more frequently.
To transmit more precise information, significantly more number of spikes would be required for the same amount of information.
Thus, in the proposed method, a trade-off exists between precision and total number of spikes.
Throughout the rest of the paper, we will refer to this method as \textit{burst coding}.

\subsection{Hybrid Neural Coding Scheme in Deep SNNs for Image Classification}
\begin{figure}[tbp]
    \centering
    \includegraphics[width=1.0\linewidth]{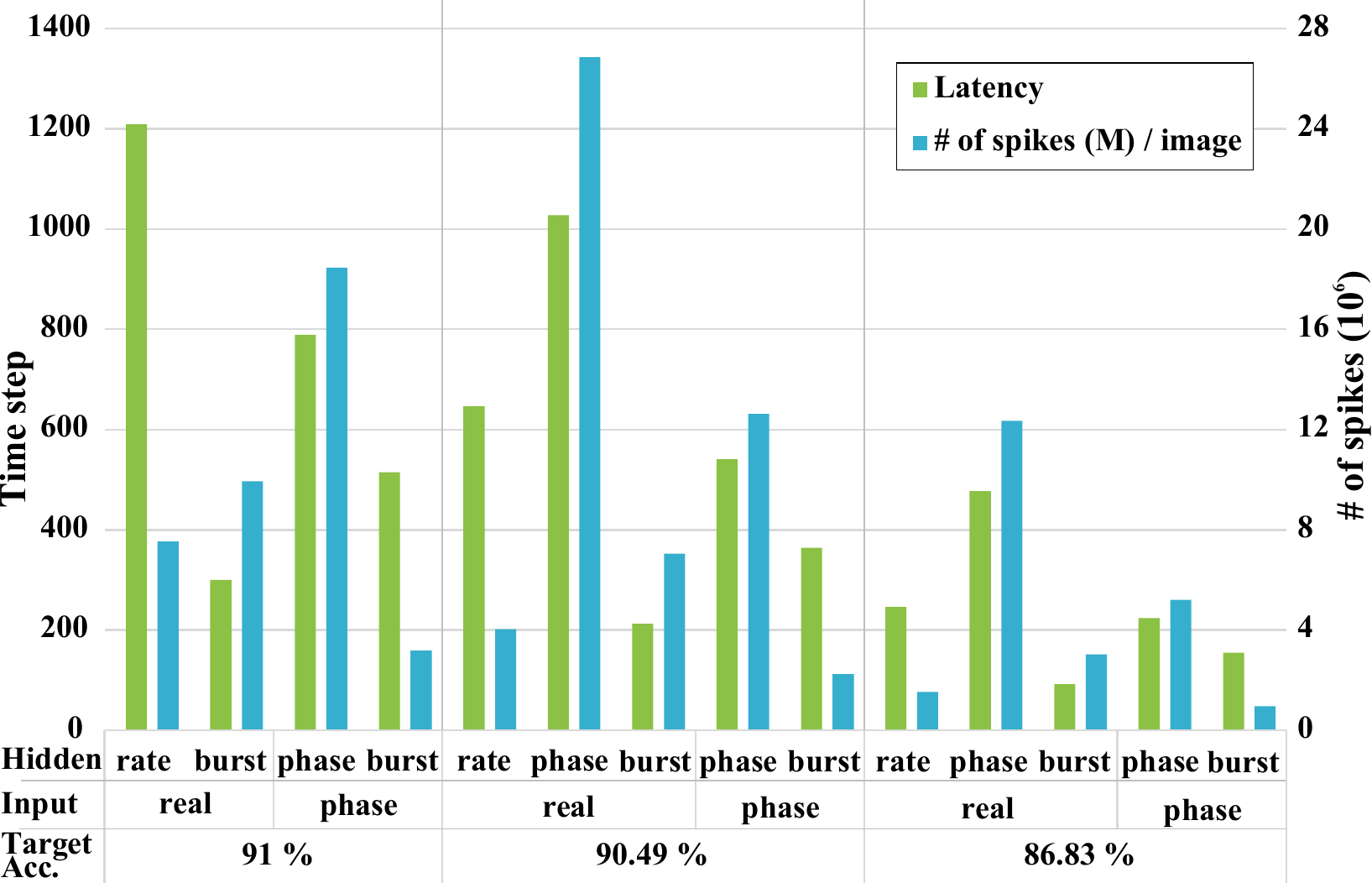}
	\vspace{-2.1em}
	\caption{Latency and number of generated spikes to reach the target accuracy of various neural coding schemes}
	\label{fig:latency_spike_target_acc}
	\vspace{-2.0em}
\end{figure}

Most of the previous studies have used only one neural coding method in the entire deep SNN~\cite{cao2015spiking,diehl2015fast,kim2018deep}. 
Only \cite{rueckauer2016theory,rueckauer2017conversion} used different neural coding schemes in input and hidden layers, but they failed to provide adequate analysis of the neural coding used in their work.
In addition, they only used real value for the input layer, which we will refer \textit{real} coding in the rest of this paper.

Using one neural coding may lead to difficulty in generating various spike patterns that are optimized for different layers in deep SNNs.
According to \cite{mochizuki2016similarity}, neurons occasionally use different neural coding schemes depending on their roles and locations in the brain.
Motivated by this observation, we propose a hybrid coding scheme that exploits different features of the layers in deep SNNs. 

To find an appropriate combination of neural coding, we divided the layers in deep SNNs into two types, namely, input and hidden layers according to layer's function.
The input layer converts continuous input values to discrete spike trains.
The input values, in many cases, are static and bounded, especially in image classification tasks.
In addition, information transmission speed and error in the input layer have a significant impact on the performance of the entire neural network.
Thus, the neural coding suitable for the input layer requires the ability to transmit static and bounded information quickly and accurately.

Considering such conditions, real or phase coding are suitable for the input layer.
Real coding is fast and accurate because it delivers the real value to the hidden layer, and phase coding can accurately convey the real value of $k$ bits in $k$ time steps.
On the other hand, rate coding needs $2^k$ time steps for $k$ bits real value, which makes the rate coding is not appropriate for the input layer.

Table~\ref{tab:exp_result_nc} describes the impact of input layer's neural coding on image classification.
Real and phase coding for input layer accomplished maximum accuracy of 91.34\% and 91.41\%, respectively.
Rate coding, however, failed to reach the DNN's accuracy in 1,500 time steps, because the input layer became a bottleneck of transmitting information.

The difference between neurons in the input layer and hidden layers lies in whether the amount of information that needs to be transferred dynamically changes.
Thus, the neural coding for hidden layers requires the capability to adaptively adjust the transmission amount.
Burst coding can dynamically determine the capacity of the transmission in an unbounded range, and thus, is suitable for hidden layers.

In our experiments, as describe in Table~\ref{tab:exp_result_nc}, the SNN model with the rate coding in hidden layer failed to reach a comparable accuracy of the DNN model in the given time steps.
Phase coding can achieve a comparable result of the DNN's result, but requires more spikes.
This disadvantage of the phase coding is illustrated in the \# of spikes column in Table~\ref{tab:exp_result_nc}.
Thus, rate and phase coding lead to degradation of efficiency in terms of latency and number of spikes generated when they are applied to the hidden layers.

\begin{figure}[tbp]
    \centering
    \includegraphics[width=1.0\linewidth]{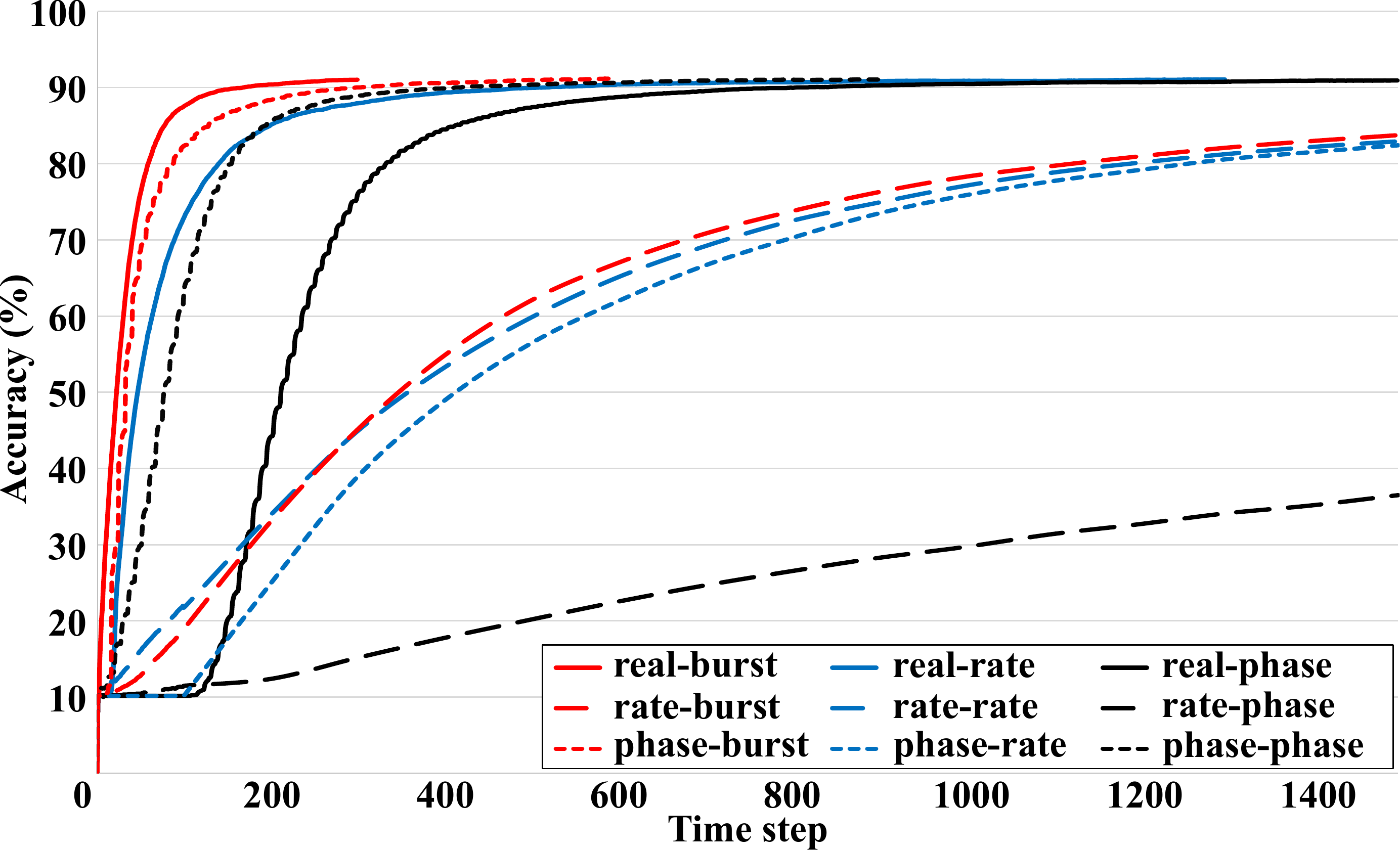}
	\vspace{-2.5em}
	\caption{Inference curve of various neural coding schemes}
	\label{fig:inference_curve}
	\vspace{-2.0em}
\end{figure}

Applying burst coding to the hidden layer yields the highest accuracy, regardless of the input layer's neural coding (the accuracy column in Table.~\ref{tab:exp_result_nc}).
This proves that the hidden neurons with burst coding scheme can generate spike patterns adaptively in order to transmit information efficiently. 

Overall, the best combination is phase coding for the input layer and burst coding for the hidden layer, which achieves the same accuracy as DNN (91.41\%) while taking less time steps and generating less spikes than other schemes.
For the rest of the paper, we use a notation, "input-hidden" to indicate neural coding of input layer and hidden layer.
For instance, phase-burst denotes that the phase and burst coding are used in input and hidden layers, respectively.

\section{Experimental Results}
%
\ctable[
star,
pos = t,
center,
caption = {Comparison of inference results with other SNN methods on various image classification datasets},
captionskip = -3ex,
label = {tab:experimental_result_other},
doinside = {\small \def\arraystretch{.7}}
]{l|cc|crcc|rrr|rr}{
    \tnote[a]{spiking density:= \# of spikes per image / (\# of neurons $\cdot$ latency)}
    \tnote[b]{normalized energy estimation results for each task}
    \tnote[c]{our experimental results}
    \vspace{-3.2em}
}{
    \toprule
    \multirow{2}{*}{Methods} & \multicolumn{2}{c|}{Neural coding} & \multirow{2}{*}{Model} & \# of & \multicolumn{2}{c|}{Accuracy (\%)} & \multirow{2}{*}{Latency} & Spikes & Spiking& \multicolumn{2}{c}{Normalized energy\tmark[b]}\\
    & Input & Hidden & & neurons & DNN & SNN & & ($10^6$) & density\tmark[a] & \footnotesize{TrueNorth~\cite{merolla2014million}} & \footnotesize{SpiNNaker~\cite{furber2014spinnaker}}\\
	\midrule
	\multicolumn{12}{l}{MNIST} \\
	\midrule
	\citeall{diehl2015fast} & rate & rate & CNN & 22,736 & 99.14 & 99.10 & 200 & 0.100 & \textbf{0.0219} & 1.000 & \textbf{1.000} \\
	\citeall{kim2018deep} & phase & phase & CNN & 22,736 & 99.20 & 99.20 & \textbf{16} & 3.000 & 8.2468 & 22.189 & 31.416 \\
	\textbf{Our Method} ($v_{\textrm{th}}$=0.125) & real & burst & CNN & 22,736 & 99.25 & 99.20 & 27 & \textbf{0.077} & 0.1245 & \textbf{0.461} & 1.061\\
	\textbf{Our Method} ($v_{\textrm{th}}$=0.125) & real & burst & CNN & 22,736 & 99.25 & \textbf{99.25} & 87 & 0.251 & 0.1270 & 0.656 & 1.613\\

	\midrule
	\multicolumn{12}{l}{CIFAR-10} \\
	\midrule
	\citeall{cao2015spiking} & rate & rate & CNN & 38,090 & 79.12 & 77.43 & 400 & 20.000 & 1.3127 & - & -\\
	\citeall{rueckauer2016theory} & real & rate & CNN & 118,282 & 87.8 & 87.8 & 280 & - & - & - & - \\
	\citeall{rueckauer2016theory}\tmark[c] & real & rate & VGG-16 & 280,586 & 91.41 & 91.06 & 1,500 & 9.334 & 0.0222 & 1.000 & 1.000 \\
	\citeall{kim2018deep} & phase & phase & CNN & 118,282 & 89.20 & 89.20 & \textbf{117} & 400.000 & 28.9038 & 143.233 & 178.708 \\
	\citeall{kim2018deep} & phase & phase & ResNet-20 & - & 91.40 & 91.40 & - & - & - & - & -\\
	\citeall{kim2018deep}\tmark[c] & phase & phase & VGG-16 & 280,586 & 91.41 & 91.21 & 1,500 & 35.196 & 0.0836 & 2.108 & 2.773\\
	\textbf{Our Method} ($v_{\textrm{th}}$=0.125) & phase & burst & VGG-16 & 280,586 & 91.41 & \textbf{91.41} & 1,125 & \textbf{6.920} & \textbf{0.0220} & \textbf{0.771} & \textbf{0.774}\\
	\textbf{Our Method} ($v_{\textrm{th}}$=0.0625) & phase & burst & VGG-16 & 280,586 & 91.41 & \textbf{91.41} & \textbf{793} & 9.342 & 0.0420 & 0.807 & 0.940\\
	
	\midrule
	\multicolumn{12}{l}{CIFAR-100} \\
	\midrule
	\citeall{kim2018deep} & phase & phase & ResNet-32 & - & 66.10 & 66.20 & - & - & - & - & -\\
	\citeall{kim2018deep}\tmark[c] & phase & phase & VGG-16 & 280,586 & 68.77 & 68.37 & 3,000 & 86.504 & 0.1028 & 1.000 & 1.000\\
	\textbf{Our Method} ($v_{\textrm{th}}$=0.125) & phase & burst & VGG-16 & 280,586 & 68.77 & \textbf{68.69} & 3,000 & \textbf{24.238} & \textbf{0.0288} & \textbf{0.712} & \textbf{0.539}\\
	
	\bottomrule
}


We evaluated various neural coding schemes using the image classification tasks, such as MNIST, CIFAR-10, and CIFAR-100, with deep SNNs.
We set accuracy, number of spikes, latency, and energy of inference as the evaluation criteria.
All the implementations are based on TensorFlow and SNN library\footnote{A public version of the code will be available.}. 

\subsection{Experimental Results for Various Neural Coding Schemes}

To assess the proposed methods, we measured the inference latency and the number of generated spikes to reach target accuracy for various configurations of neural coding.
In these experiments, we trained VGG-16 as a baseline DNN model on the CIFAR-10 dataset (test accuracy: 91.41\%), and then converted the trained DNN model to SNN.
We set three types of the target accuracy: 91.0\%, 90.49\%, and 86.83\%.

The experimental results are presented in Fig.~\ref{fig:latency_spike_target_acc}.
It should be noted that we excluded the results of some neural coding schemes from the graph because they failed to reach the target accuracy in 1,500 time steps.
When using rate coding in the input layer, it failed to achieve all the target accuracy within the given time step.
For the all targets, the real-burst reached the target the fastest, and the phase-burst had the least number of spikes to reach target accuracy. 

In terms of latency, the burst coding was the fastest method in the hidden layer regardless of the neural coding of the input layer.
From the aspect of number of spikes, in all cases, the most spikes occurred when phase coding was applied to the hidden layer.
the rate and burst coding generated the least spikes when we used the real and phase coding in the input layer, respectively.
However, the real-rate configuration increased the latency rapidly as the target accuracy increased.
This can cause significant performance degradation if the SNN model require high accuracy.

The inference curves for various configurations are illustrated in Fig.~\ref{fig:inference_curve}.
Deep SNNs with rate coding in the input layer showed slower convergence speed than the others, and burst coding, in the hidden layer, yielded rapid convergence speed.
In the case of rate-phase, the inference curve showed worst performance among the configurations.
It means that the phase coding has less adaptability than other neural coding.
On the contrary, the burst coding showed high adaptability and performance in hybrid neural coding scheme.
The experimental results in Fig.~\ref{fig:latency_spike_target_acc} and \ref{fig:inference_curve} led us to posit that the burst coding and hybrid coding scheme empower deep SNNs in many aspects, namely, speed, accuracy, and efficiency.


\subsection{Comparison with Other Deep SNN Algorithms}
To validate our proposed methods, we compared them with previous works on MNIST, CIFAR-10, and CIFAR-100 datasets in terms of classification accuracy, latency, number of spikes, and energy consumption as shown in Table~\ref{tab:experimental_result_other}.

Regardless of which neural coding scheme is used, the SNN achieved comparable accuracy with that of DNN for the MNIST dataset.
In the case of CIFAR-10 and -100 datasets, we could obtain the accuracy close to the DNN when applying phase coding or our proposed methods.
Overall phase coding showed fast inference speed.
However, the accuracy given in \cite{kim2018deep} is lower than the accuracy of the baseline DNNs we used, or there is no information on latency and number of spikes.
Thus, we applied phase coding to our model for fair comparison.
As a result, the deep SNN model with phase coding cannot reach to the baseline DNN's accuracy within the given time step (1,500) for the CIFAR-10 dataset.
On the contrary, the deep SNN model with our proposed methods achieved the accuracy.
Experimental results with different $v\textrm{th}$ showed the effect of the precision in the burst coding.
The higher the precision (smaller $v\textrm{th}$), the faster and more accurate inference is possible, but the more spikes occur.

The number of spikes generated using phase coding was greater than that obtained using any other method, which has also been shown by \cite{kim2018deep}.
However, as the latency increases, the number of spike occurrences increases, so we proposed spiking density for a fair comparison.
Spiking density is the expected value of the number of spikes generated in a neuron per time step, and means that the lower the value, the more sparse spikes occur.
The result indicated that burst coding has the lowest spiking density of 0.0220, which is considerably low compared to others.
In the our experiment with the method of \cite{rueckauer2016theory}, the spiking density (0.0222) was similar to that of the proposed method, but the latency was much longer and the accuracy was lower than those of our method.

We estimated energy consumption from two neuromorphic architectures (TrueNorth~\cite{merolla2014million} and SpiNNaker~\cite{furber2014spinnaker}) to investigate the effect of the proposed methods on the energy efficiency of deep SNNs.
The energy estimates were obtained by dividing the total energy consumption by computation, routing, and static energy, and then calculating each energy proportionally to the number of spikes, spiking density, and latency, respectively.
We referred to \cite{merolla2014million,furber2014spinnaker,moradi2018impact} for the energy consumption ratios of the three parts (computation, routing, and static), and the estimated values were normalized for each dataset. 
As a result of our estimation, our method consumed the least energy in almost all cases with higher accuracy.
Thus, the proposed methods enable deep SNNs to transmit information fast and efficiently.

\section{Discussion}

\begin{figure}[tbp]
    \centering
    \includegraphics[width=0.9\linewidth]{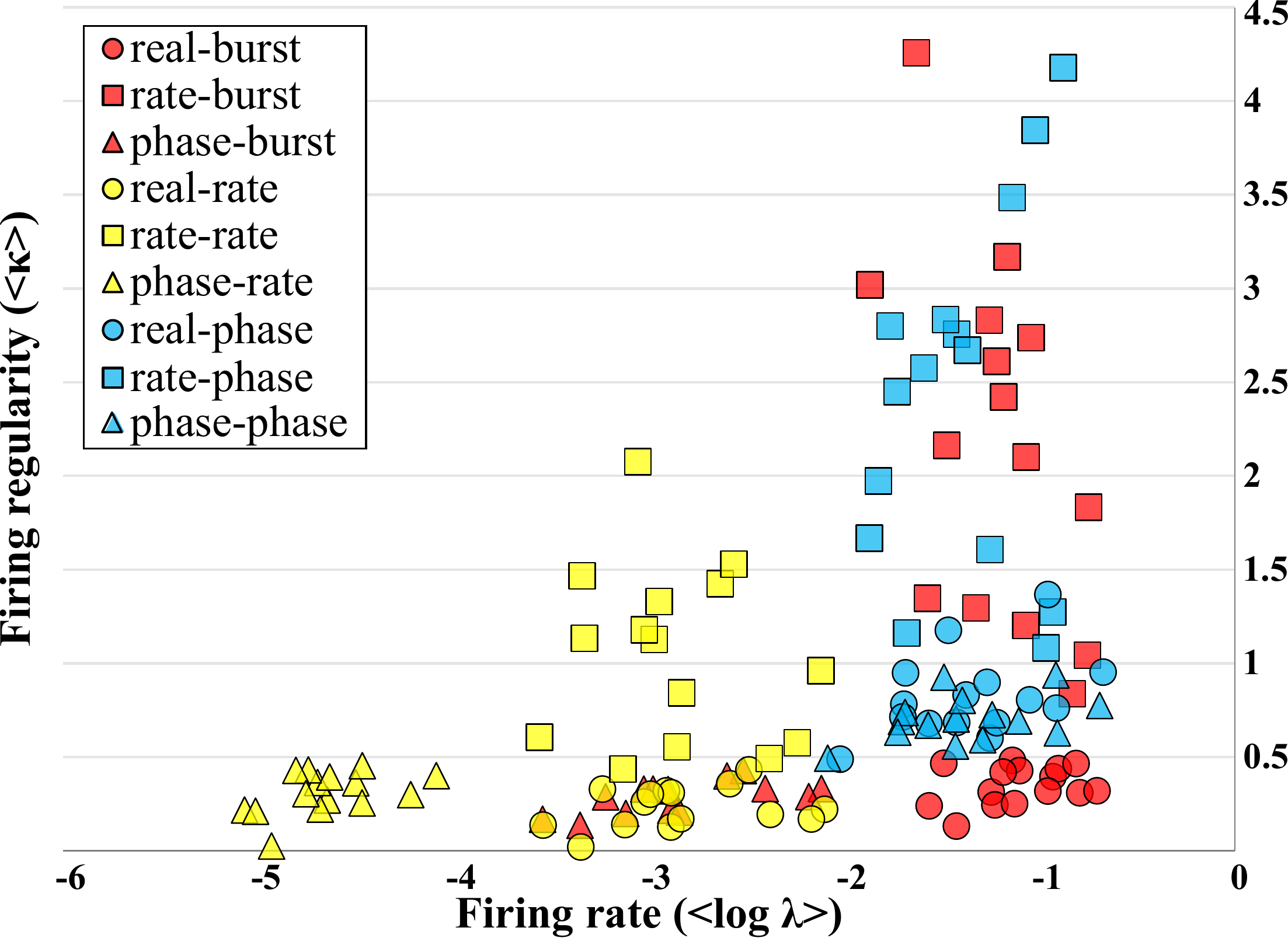}
    \vspace{-1.5em}
	\caption{Firing rate-regularity graph for various neural coding schemes}
	\label{fig:firing_characteristics}
    \vspace{-1.5em}
\end{figure}

The neural spike pattern analysis typically aims to analyze neurons' characteristic and their neural coding scheme in depth~\cite{mochizuki2016similarity}.
To analyze spike patterns of the proposed burst coding and hybrid neural coding scheme, firing characteristics were determined in our experiments.
We calculated firing rate and regularity of spike trains as the firing characteristics for each neural coding scheme.
The spike trains were measured from randomly sampled 10\% of the neurons in each layers for 10,000 time steps.

The firing rate of a neuron $\lambda$ can be defined as
\begin{equation}
\label{eq:firing_rate}
    \lambda = \frac{n}{\sum_{i=1}^{n} I_i} \textrm{,}
\end{equation}
where ${I_i}$ is the duration of the ${i}$th ISI, and ${n}$ is the number of ISIs in a spike train.
Firing regularity is defined by the regularity of the spike train generated.
We used the coefficient of variation to obtain the firing regularity ${\kappa}$ in a spike train as follows:
\begin{equation}
\label{eq:firing_regularity}
    \kappa = \frac{std(I)}{mean(I)} \textrm{.}
\end{equation}
We averaged log firing rate < log$\lambda$ > and firing regularity < $\kappa$ > for the spike pattern analysis.
The firing rate-regularity graph for various neural coding schemes is shown in Fig.~\ref{fig:firing_characteristics} .
The shapes and colors of each dot indicate various neural coding schemes of the input and hidden layers, respectively.

When analyzing the characteristics of the colored clusters, we found that the phase coding in the hidden layer (blue dots) has the highest firing rate regardless of the input layer's neural coding.
This illustrates the low flexibility of phase coding.
Thus, the phase coding is not suitable in the hidden layer for the hybrid neural coding method.

On the contrary, the rate coding in the hidden layer (yellow dots) can lead to flexibility depending on the coding scheme in the input layer.
However, it is not able to transmit information quickly at a low firing rate, such that classification latency may be longer than that for other neural coding schemes. 

The burst coding in the hidden layer (red dots) tends to have the largest flexibility depending on the coding method in the input layer. 
This means that burst coding is well-adapted to different coding schemes in the input layer.
Thus, it is a suitable coding scheme for the hidden layer of the hybrid neural coding scheme.

\section{Conclusion}
In this paper, we analyzed the effect of various neural coding schemes in deep SNNs.
Our analysis revealed that rate coding lacks the ability to transmit information efficiently while phase coding has low computational energy-efficiency.
Based on our analysis, we propose burst coding and hybrid neural coding scheme, which are inspired by human brain, to improve energy-efficiency and speed of inference in deep SNNs.
The proposed methods not only achieve reasonable accuracy, but also substantially improve inference efficiency in terms of speed and energy.
We expect that the proposed methods can be the applied in deep SNNs for other tasks to improve the efficiency.

\bibliographystyle{IEEEtranN}
\bibliography{dac2019}

\end{document}